\documentclass{article}

     \usepackage[final]{neurips_2019}

\usepackage[utf8]{inputenc} 
\usepackage[T1]{fontenc}    
\usepackage{hyperref}       
\usepackage{url}            
\usepackage{booktabs}       
\usepackage{amsfonts}       
\usepackage{nicefrac}       
\usepackage{microtype}      
\usepackage{graphicx}
\usepackage{caption}
\usepackage{subcaption}

\title{Waymo Driverless Car Data Analysis and Driving Modeling using CNN and LSTM}

\author{%
  Aashish Kumar Misraa \\
  Columbia Universtiy\\
  \And
     Naman Jain \\
     Columbia Universtiy \\
  \And
     Saurav Singh Dhakad \\
     Columbia Universtiy \\
}

\makeatletter
\newcommand{\@notice}{}
\makeatother

\begin{document}

\maketitle

\begin{abstract}
Self driving cars has been the biggest innovation in the automotive industry, but to achieve human level accuracy or near human level accuracy is the biggest challenge that research scientists are facing today. Unlike humans autonomous vehicles do not work on instincts rather they make a decision based on the training data that has been fed to them using machine learning models using which they can make decisions in different conditions they face in the real world. With the advancements in machine learning especially deep learning the self driving car research skyrocketed. In this project we have presented multiple ways to predict acceleration of the autonomous vehicle using Waymo's open dataset. Our main approach was to using CNN to mimic human action and LSTM to treat this as a time series problem.
\end{abstract}

\section{Introduction}

CNN have revolutionized the pattern recognition and computer vision fields by surpassing tradition computer vision and machine learning approaches. This had a direct impact on self driving research. And indeed, many researchers started using CNN and its variants in their self driving research.

Neural networks are inspired by how human brain works. It is a replica of human nervous system. A simple neural network can take several inputs and assign them weights and then sums them up and pass through an activation function to calculate the output. When we add convolutional layer to this neural network we get a Convolutional Neural Network. A Convolution layer is usually used with a pooling layer. Convolutional layer helps the network to learn different features of the image with the help of a filter. These features can be edges, boundaries etc. We have used CNN to predict acceleration of the AV using the camera images because there is a direct correlation of what a driver sees and the acceleration of the car and to exploit this correlation we have used CNN.

Since we wanted to treat this problem of acceleration prediction we also used LSTM in our advanced model since we believe that your future actions depends on your past actions even in the  case of driving  a car. Therefore, using LSTM was the obvious choice. Long short term memory also known as LSTMs have the capability to learn long term dependencies. As you can see in Figure 1 having a chain like structure helps LSTM to not only used the information of the current state but also from previous state.

Therefore, we have used different combinations of CNN and LSTM to predict the acceleration of the autonomous vehicle which will be discussed further. 

\begin{figure}
  \centering
  \includegraphics[width=0.7\linewidth]{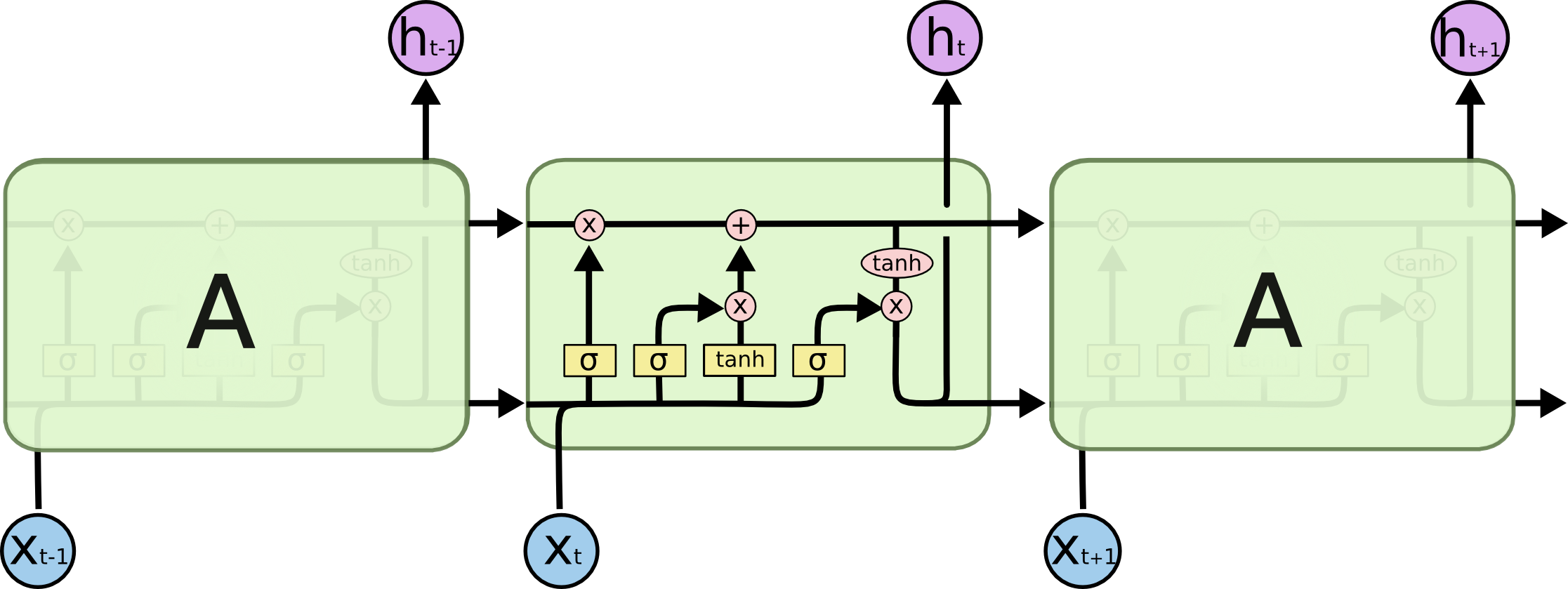}
  \caption{Basic structure of LSTM}
\end{figure}

\subsection{Behavioural Cloning}

Behavioural cloning is an approach where you mimic the actions of the driver. Mariusz Bojarski et al [1] from NVIDIA presented an approach where they used to CNN to mimic the driver and predicted steering commands.They trained a CNN to map raw pixels from a single front facing camera to steering commands. The results were surprisingly good as to predict steering commands accurately simply from images was never done before.This system was able to work even in areas where the vision was not clear and even on unpaved roads which is why it proved to be surprisingly powerful.

In figure 2, which shows the network for the training the system. In this system CNN is trained using images as input which then computes the steering command for the autonomous vehicle. The weights are adjusted using the backpropgation by computing error, which tries to compute the CNN output as close as ground truth. 

\begin{figure}
  \centering
  \includegraphics[width=0.8\linewidth]{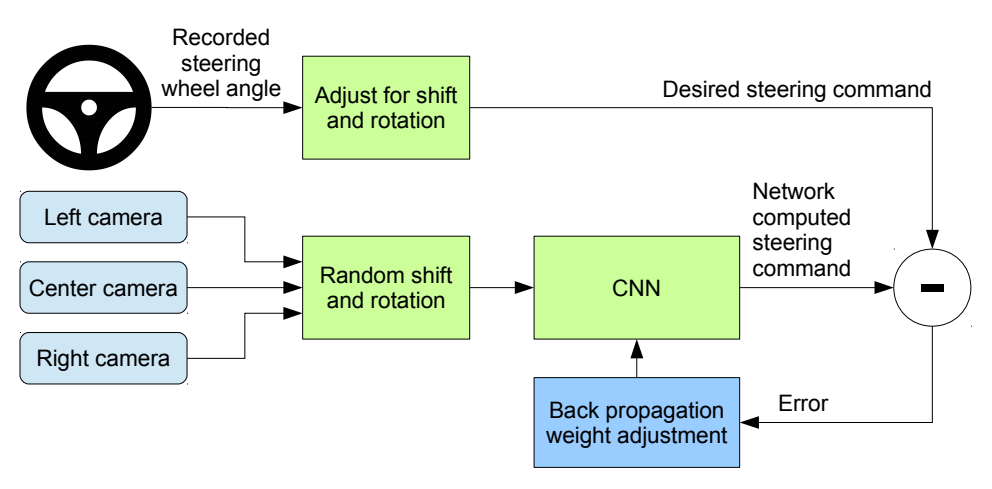}
  \caption{Proposed network of NVIDIA paper}
\end{figure}

\subsubsection{Network Architecture}

Mariusz Bojarski et al [1] used 9 layer architecture which includes including a normalization layer, 5 convolutional layers and 3 fully connected layers. The first layer of the plane is responsible for the normalization of the input image. Following the normalized input they have 5 convolutional layers which works as feature extractor and then 3 dense layer which provides vehicle control as output.

\begin{figure}
  \centering
  \includegraphics[height=0.8\linewidth]{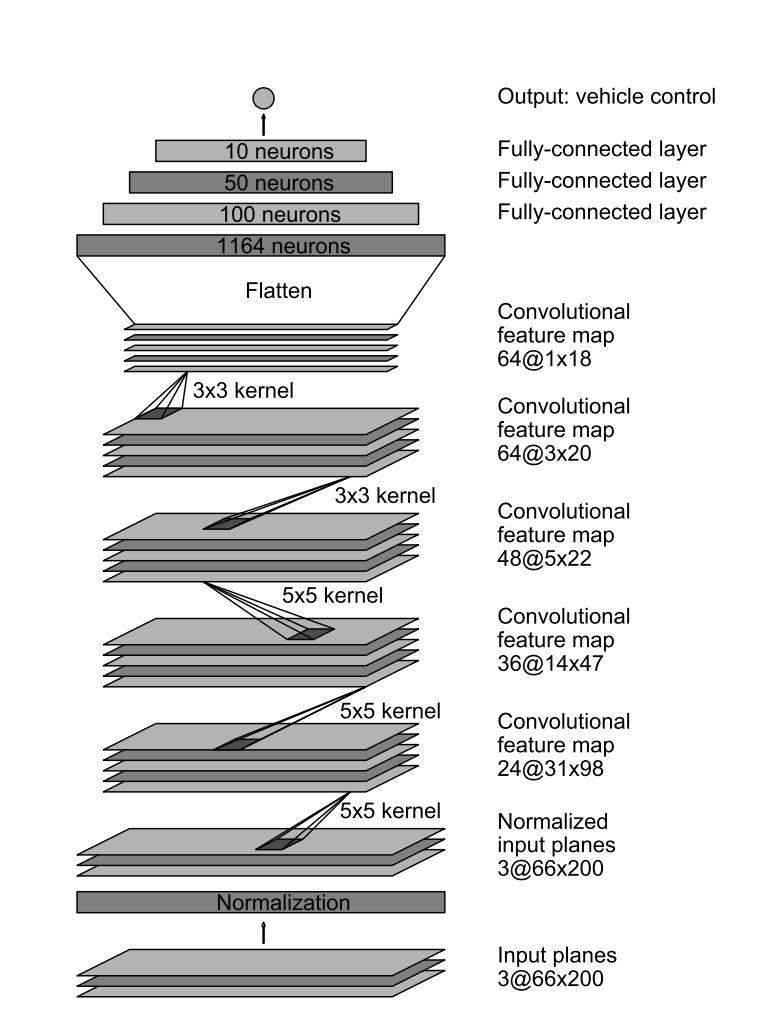}
  \caption{Network Architecture of NVIDIA paper}
\end{figure}

\subsection{Other Related works}

Since predicting the acceleration of self driving cars can be perceived as a time series problem, we came across different researches that involved a combination of convolution and time series. Andrej Karpathy et al [2] presented a model that generated natural language descriptions of images and their regions. Their multi modal approach helped them to generate sentence description of the data by exploiting the audio-visual correlation. We used this as a motivation for computing acceleration of AV using waymo dataset in out project.

\subsection{Organization}

The organization of the paper is  as follows. In Section 2, we have defined the problem and assumptions related to the project. In section 3 we have explained the waymo dataset that we used for this project. In section 4 we have discussed about the feature extraction and prepossessing that we did in this project. Section 5 explain the different models used in this project. At last, in section 6 and 7 we have results and conclusion as well as future works that needs to be done.

\section{Problem Specification}

The main goal of this project was to infer how the waymo's self driving car works. For this we used waymo's self driving car dataset which they released recently. Though the dataset have most of the information about their autonomous vehicle but the dataset does not explicitly tell anything about how their algorithm works. Since their algorithm is not open sourced, our primary goal was to infer their algorithm. Also, the dataset is not designed for traffic so we needed to determine that how would it work in different scenarios such as traffic etc.

\section{Dataset}

The most important part in any machine learning task is data. Over 10 millions miles data have been collected during last few years in over 25  cities. This data is crucial in the advancement of the self driving cars.
Waymo announced their dataset publicly to help advance this research.
This dataset is a high quality high-quality multi-modal sensor dataset for autonomous driving which is available for free for researchers. It is a combination of high resolution camera and sensor data collected via waymo's autonomous vehicle.
The dataset has a variety of scenarios from compact urban areas to suburban landscapes. It also consists of data collected during different weather conditions during day and night.

It contains a total of 1000 driving segments. Here each segment is of 20 seconds of driving data collected by AV. Each segment is comprised of 200,000 frames captures at 10 Hz per sensor. This is a very rich data to work upon.

There were a total of  five high-resolution Waymo lidars and five front-and-side-facing cameras used to collect this dataset. This gives us dense labelled data of pedestrians, vehicles etc. with camera lidar synchronization. There are a total of 12 million 3D labels and 1.2 millions 2D label.
In this project we haven't used any label information as we worked on the raw images captured by the camera.

Also, we used waymo's open dataset training data for training and validation data for testing for our all models.

\section{Feature Extraction and Preprocssing}

Waymo dataset do not have the acceleration of the autonomous vehicle. But it did had the velocity of the AV in each frame, using this information we computed ground truth acceleration for each frame. Acceleration of a frame is nothing but difference in velocity of the current frame and the previous frame.

\[ a_{curr} =  v_{curr} - v_{prev} \]

Waymo dataset have following information given corresponding to each frame: velocity of AV in x, y and z axis, velocity of a vehicle in front of AV in x, y and z axis if there is one present. Using the ground truth acceleration we were able to able a 11 feature set corresponding to each frame. We set the velocity of front vehicle to be 0 if there was none present. Feature set had the following features: \emph{vx, vy, vz, dx, dy, vfx, vfy, vfz, afx, afy, afz} which are velocity and acceleration of AV and the vehicle in front of it in x, y and z axis and the relative distance of AV in x and y direction.

Since, we used CNN we have used the images of front camera as our training input. Since the image size we too large, we had to down-sample it to a size if xxxx so that our training time would decrease.

\section{Models}

We tried different combinations of models to predict acceleration of the autonomous vehicle using the 11 feature set, images and the time data that we had.

\subsection{Baseline (NN) Model}

Our baseline model was just a basic deep neural network. As we can see in Figure 4, for our baseline we had a deep neural network with 11 feature set as our input and 2 hidden layer with 8 and 4 neurons respectively. To see complete architecture, please visit: \url{https://github.com/toshi1801/team4/blob/master/model_architectures/nn.png}

\begin{figure}
  \centering
  \includegraphics[width=0.6\linewidth]{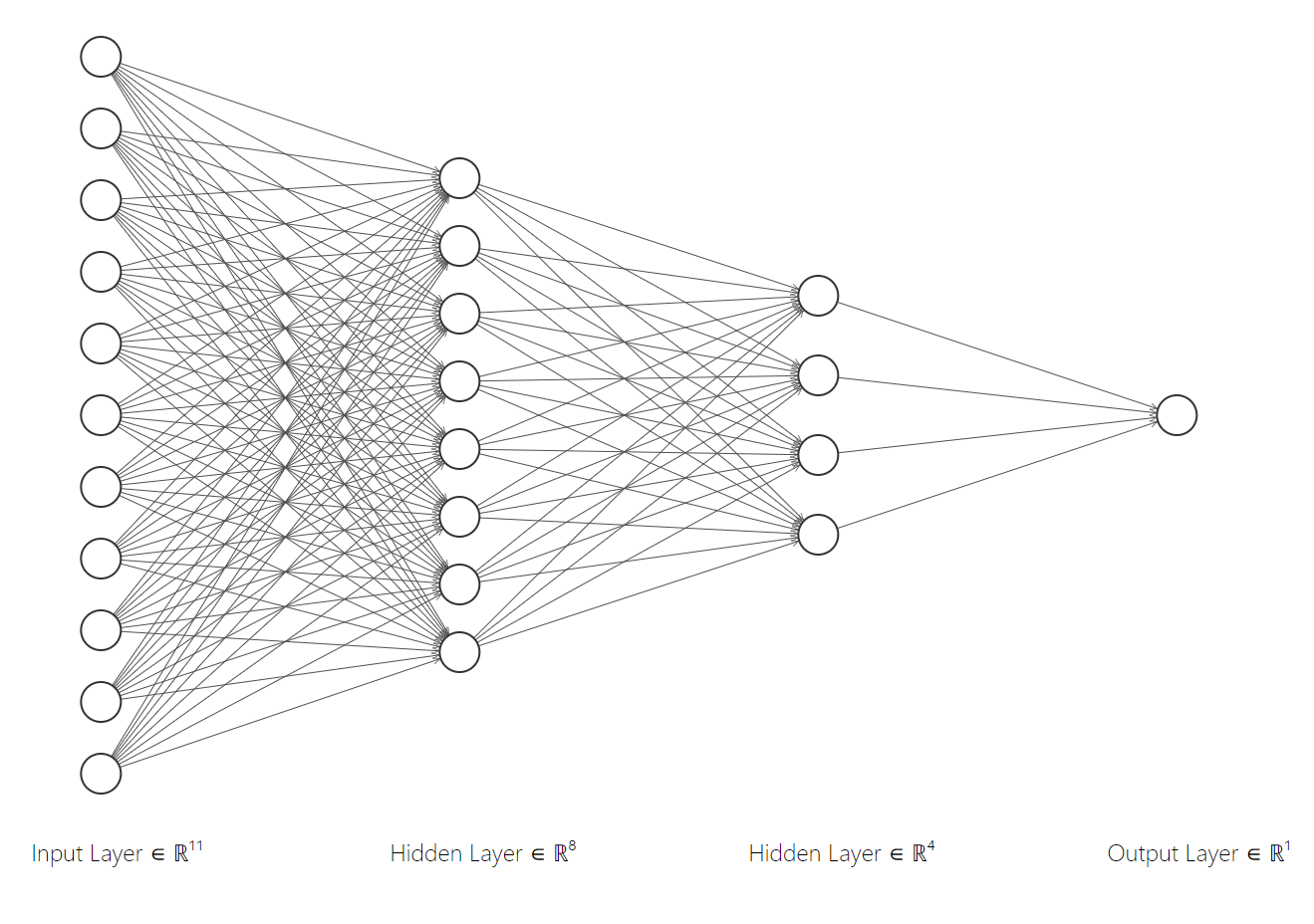}
  \caption{Network Architecture of Baseline Model}
\end{figure}

\subsection{CNN Model}

Next we tried a Convolutional Neural Network with number of Convolutional, Batch Normalization and Pooling layers. Here we used images of each frame as an input to predict acceleration of each frame. As we can see in Figure 5, the output of Convolutional, Batch Normalization and Pooling layers is then fed into some dense layers to extract feature embedding of the image and then predict the output. To see complete architecture, please visit: \url{https://github.com/toshi1801/team4/blob/master/model_architectures/cnn.png}

\begin{figure}
  \centering
  \includegraphics[width=0.9\linewidth]{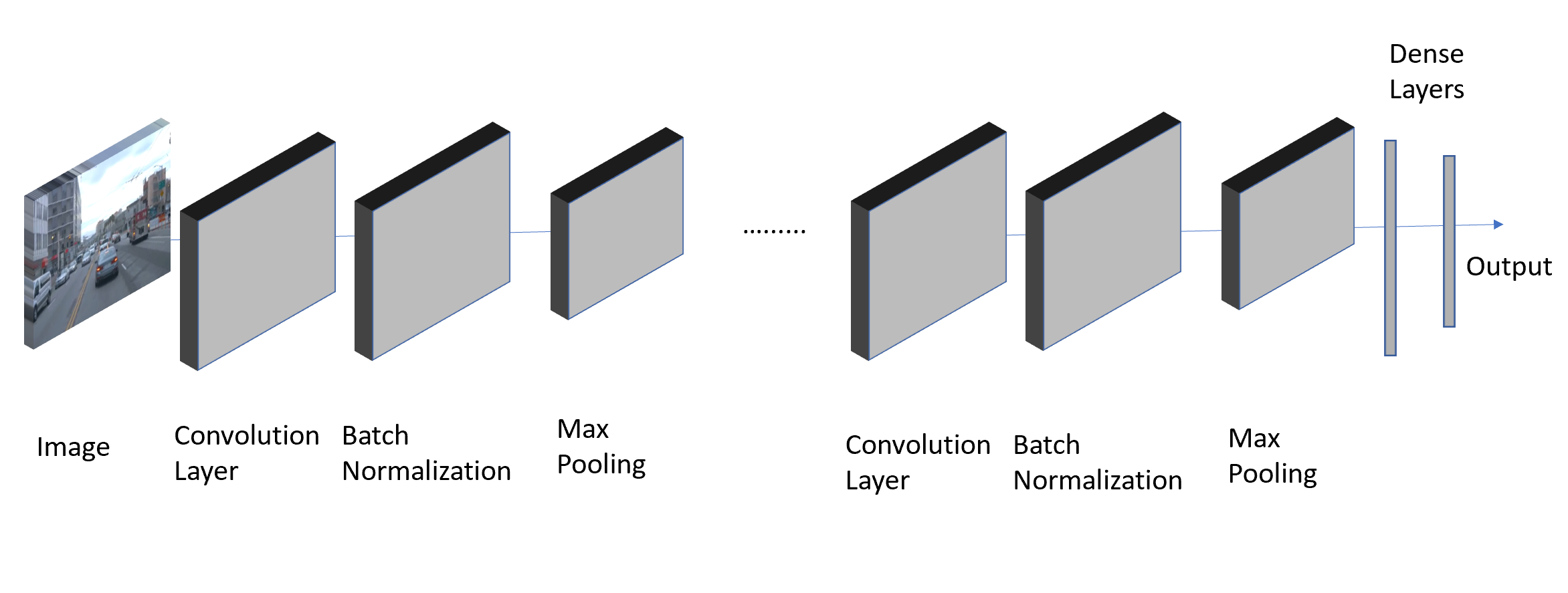}
  \caption{Network Architecture of CNN Model}
\end{figure}

\subsection{CNN+NN Model}

After trying CNN and NN model individually, we decided to merge those two models to better represent a frame using the image embedding as well as the 11 feature corresponding to a frame. This helped us increasing the accuracy of the prediction. For this model, we took the vanilla CNN model and NN model used previously and concatenated the output of dense layers from CNN model and the dense layer output of the NN model and then again fed them into a couple of dense layers to predict the output. To see complete architecture, please visit: \url{https://github.com/toshi1801/team4/blob/master/model_architectures/cnn_nn.png}

\begin{figure}
  \centering
  \includegraphics[width=0.9\linewidth]{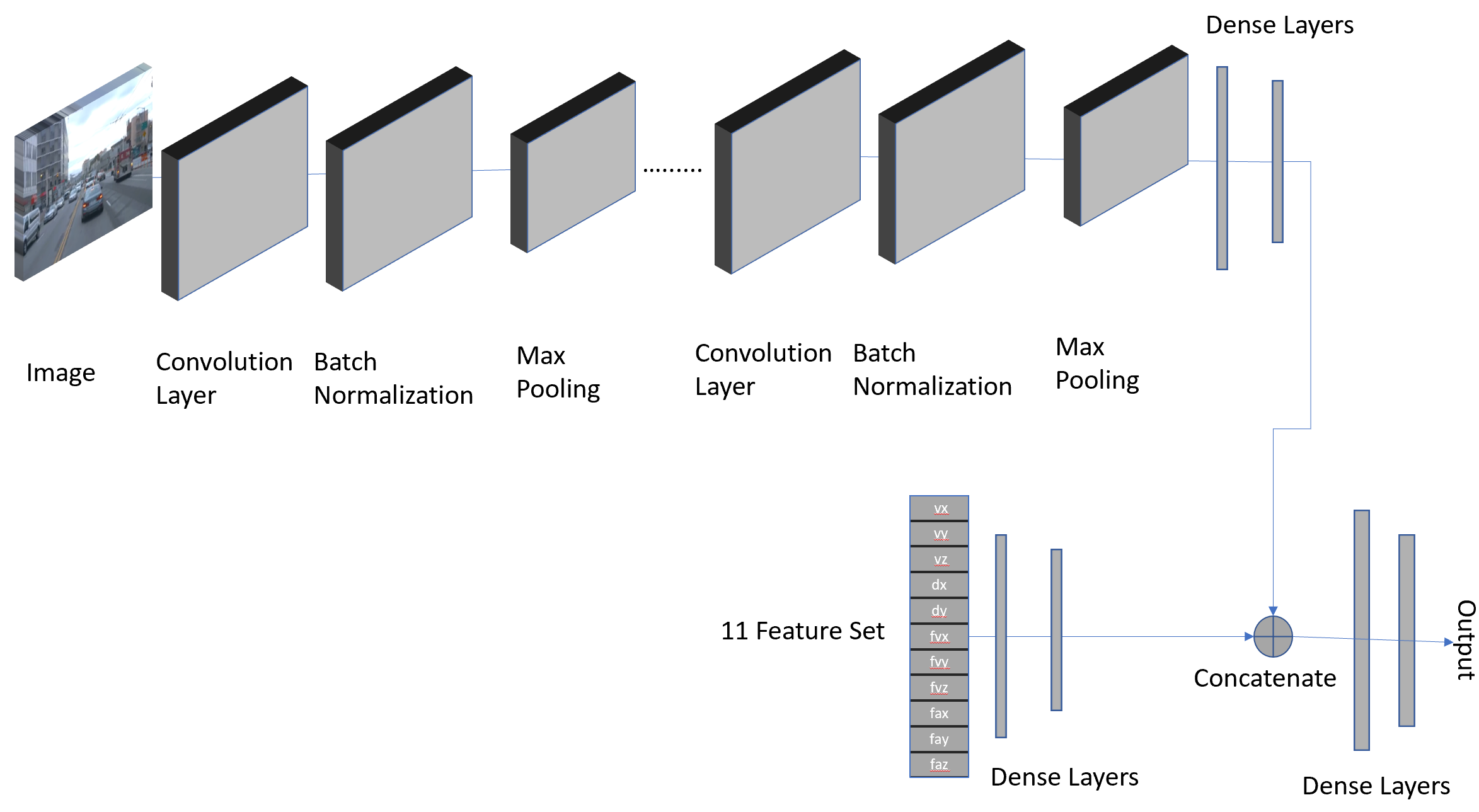}
  \caption{Network Architecture of CNN+NN Model}
\end{figure}

\subsection{CNN+LSTM}

Since using only current frame image to predict the acceleration is not giving any information of the past. Therefore, we implemented CNN with LSTM to get information not only from the current frame but from past 5 frames.For a particular frame we calculated image embeddings of the current frame as well as last 4 frames and then fed this embeddings to the LSTM and time distributed dense layer to predict the acceleration. Figure 7 illustrates the CNN+LSTM model. To see complete architecture, please visit: \url{https://github.com/toshi1801/team4/blob/master/model_architectures/cnn_lstm.png}

\begin{figure}
  \centering
  \includegraphics[width=0.9\linewidth]{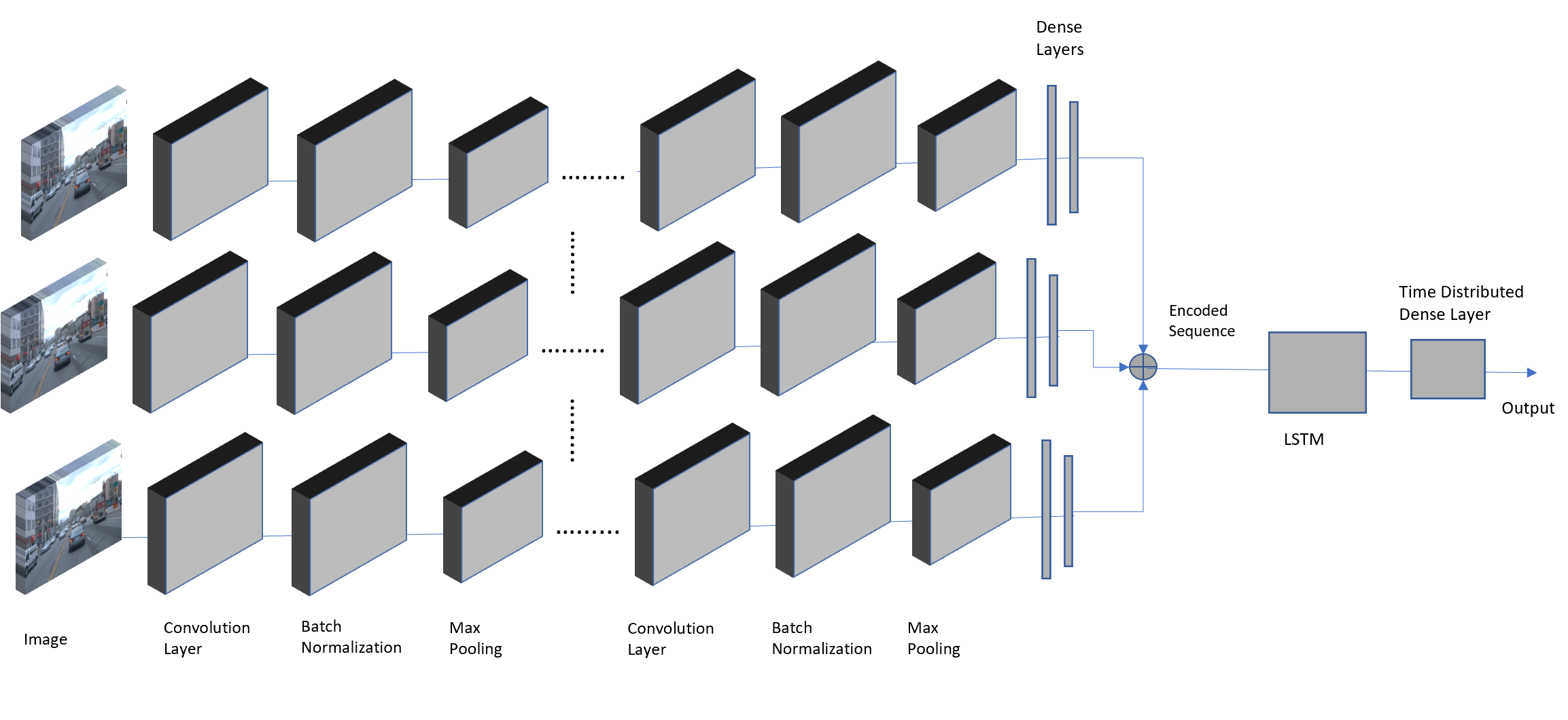}
  \caption{Network Architecture of CNN+LSTM Model}
\end{figure}

\subsection{Advanced Model}
For our advanced model we combined the idea of all previous models into one. We used transfer learning in this model to better extract features of the frame image. We used CNN model as our base model to serve as feature extractor then fed the output of CNN through some dense layer to get the image embeddings. Then, as we did in CNN+NN model we concatenated this image embeddings with the output of 11 feature set from a couple of dense layer then passed this finally to LSTM and time distributed dense layer which can be seen in Figure 8. This model outperformed every other model proposed. To see complete architecture, please visit: \url{https://github.com/toshi1801/team4/blob/master/model_architectures/advanced.png}

\subsubsection{Time Distributed Dense Layer}
Time Distributed dense layer is used mainly in RNNs including LSTM, to keep one-to-one relations on input and output. Assume you have 60 time steps with 100 samples of data (60 x 100 in another word) and you want to use RNN with output of 200. If you don't use Time Distributed dense layer, you will get 100 x 60 x 200 tensor. So you have the output flattened with each timestep mixed. If you apply the Time Distributed dense, you are going to apply fully connected dense on each time step and get output separately by time steps.

\begin{figure}
  \centering
  \includegraphics[width=0.9\linewidth]{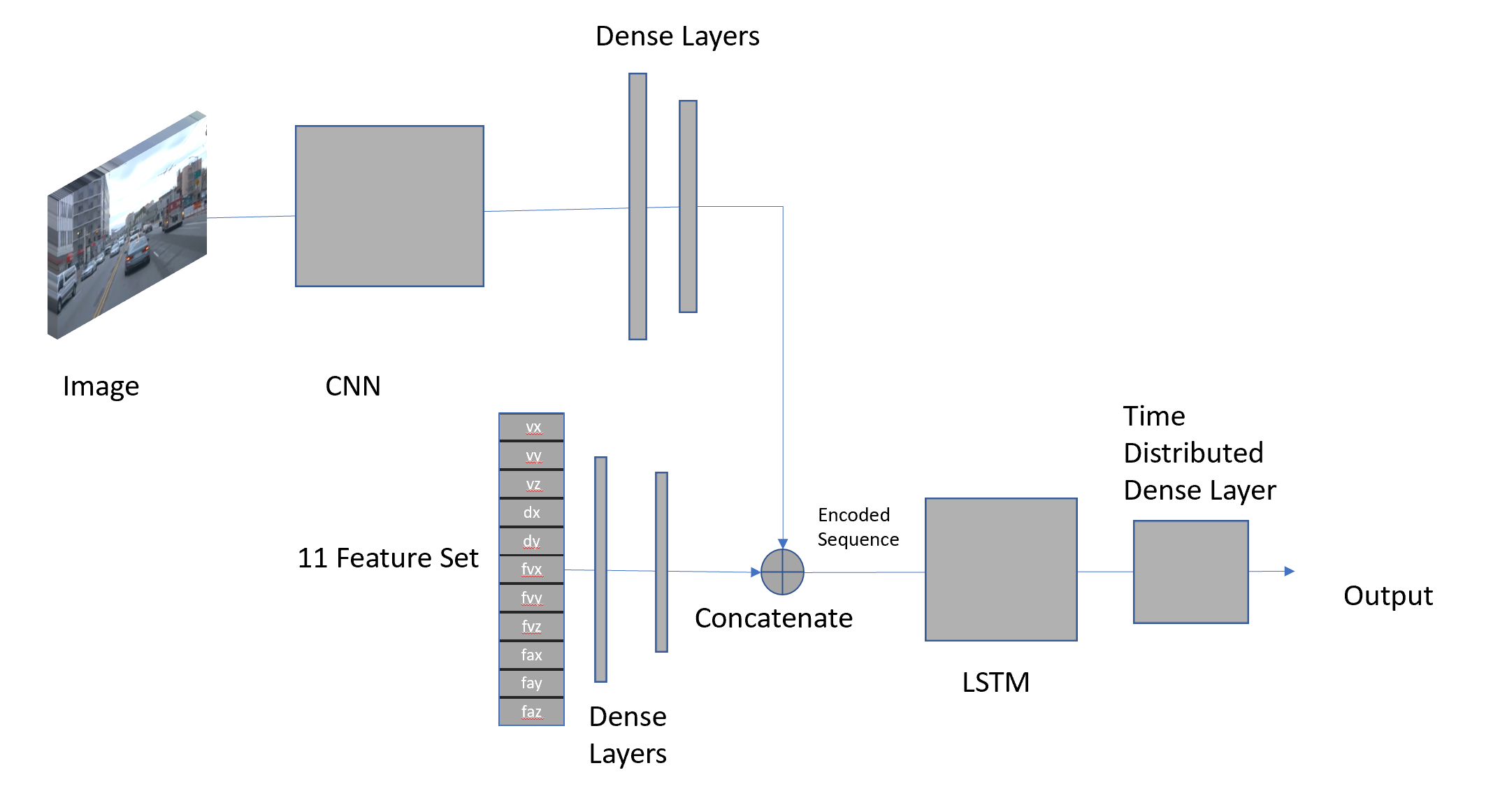}
  \caption{Network Architecture of Advanced Model}
\end{figure}

\section{Experiments and Results}

\subsection{Model Loss}

In every machine learning algorithms we try to either maximize or minimize a function objective function. In order to acheive this we use a function called loss function, which tells us how good our machine learning algorithm is performing or how close is our prediction to the true value. Loss function can be classified mainly as of two types: Regression loss and Classification loss.

There are various loss function that can be used for different problems and situation. Since our problem was to predict the output of the AV we have used MAE (Mean Absolute Error) as our loss function which is a type of regression loss function. It is basically the sum of absolute differences between our target and predicted value. 

\[ MAE = \frac{\sum_{i=1}^{n} |y_{i} - y_{i}^{p}|}{n}  \]

\subsection{Data Size}

Data size plays a crucial role in any deep learning model. We experimented with different data size. We used 1 tar (25 segments) as well 10 tar (250 segments) to train our models. Since smaller data size helped us train quickly we were able to reiterate process quickly and make necessary changes. For our final model we used the complete waymo dataset.

\subsection{Model Loss Plots and Training Time}

Figure 9 shows the plots of model loss of different models that we discussed above. There are a few key things to infer from these plots. Firstly, the model loss decreased when we combined CNN and NN which used both the 11 features and images in comparison to the NN with 11 features only or CNN with images only model. That means, that though the calculated 11 features are important in determining the acceleration of the AV, the images also plays an important factor. Secondly, in every model that used CNN we can see that the model loss decreased with the increase in training data that is when we used 1 tar the model loss was higher in comparison to the model in which we used 10 tars. Lastly, the advanced model (see Figure 10) outperformed every other model which indicates the use of LSTM with CNN and NN is a correct approach in this kind of problem. Also, we used the whole training set for this model which also played an important role in lowering the overall loss.

\begin{figure}
     \centering
     \begin{subfigure}[b]{0.3\textwidth}
         \centering
         \includegraphics[width=\textwidth]{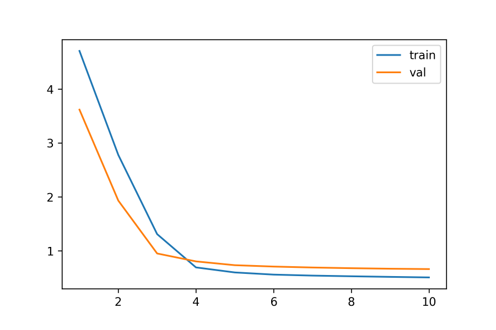}
         \caption{Baseline Model (1 tar)}
         \label{fig:Baseline Model (1 tar)}
     \end{subfigure}
     \hfill
     \begin{subfigure}[b]{0.3\textwidth}
         \centering
         \includegraphics[width=\textwidth]{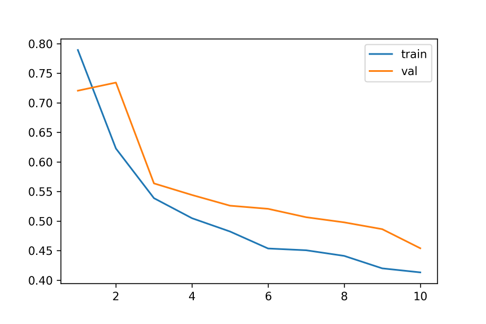}
         \caption{CNN Model (1 tar)}
         \label{fig:CNN Model (1 tar)}
     \end{subfigure}
     \hfill
     \begin{subfigure}[b]{0.3\textwidth}
         \centering
         \includegraphics[width=0.8\textwidth]{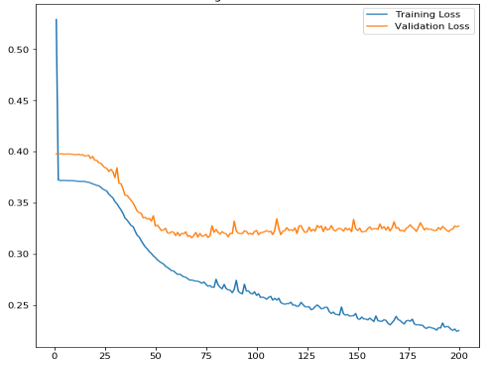}
         \caption{CNN Model (10 tar)}
         \label{fig:CNN Model (10 tar)}
     \end{subfigure}
     \hfill
    \begin{subfigure}[b]{0.3\textwidth}
         \centering
         \includegraphics[width=\textwidth]{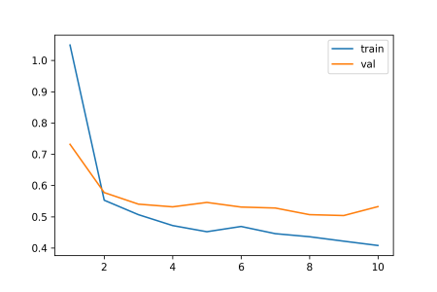}
         \caption{CNN+NN Model (1 tar)}
         \label{fig:CNN+NN Model (1 tar)}
     \end{subfigure}
     \hfill
     \begin{subfigure}[b]{0.3\textwidth}
         \centering
         \includegraphics[width=0.8\textwidth]{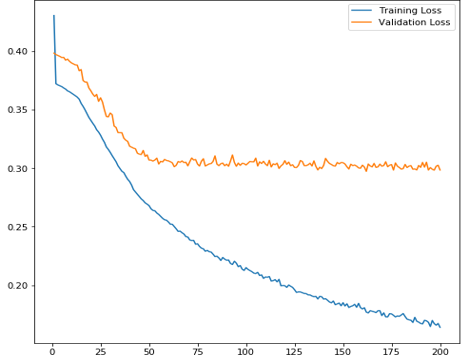}
         \caption{CNN+NN Model (10 tar)}
         \label{fig:CNN+NN Model (10 tar)}
     \end{subfigure}
     \hfill
     \begin{subfigure}[b]{0.3\textwidth}
         \centering
         \includegraphics[width=\textwidth]{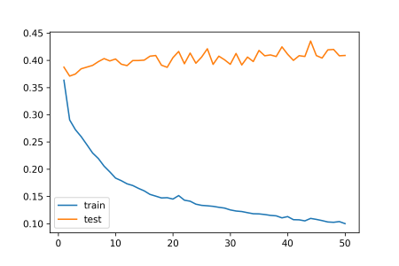}
         \caption{CNN+LSTM Model (10 tar)}
         \label{fig:CNN+LSTM Model (10 tar)}
     \end{subfigure}
        \caption{Model Loss Plots}
        \label{fig:three graphs}
\end{figure}

\begin{figure}
  \centering
  \includegraphics[height=0.5\linewidth]{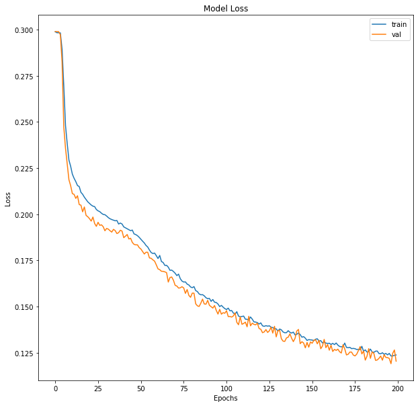}
  \caption{Advanced Model Loss over whole dataset}
\end{figure}

Table 1 shows the model loss on training as well as validation data for every model as well the time required to train each model. The time was likely to increase with the increase in size of images or dataset and that is what happened.

\begin{table}
  \caption{Results}
  \label{sample-table}
  \centering
  \begin{tabular}{lllll}
    \toprule
    \cmidrule(r){1-2}
    Model     & Data Size (No. of tar file)     & Train Loss    &Val Loss     &Train Time \\
    \midrule
    Baseline & 1    & 0.5086 & 0.6634 & 3 mins    \\ 
    Baseline     & 10 &  0.3415 &  0.4014 & 10 mins    \\ 
    CNN & 1    & 0.4133 & 0.4541 &   15 mins  \\   
    CNN     & 10 &  0.2251 & 0.3272 & 2 hrs 45 mins     \\ 
    CNN+NN & 1    & 0.1080 & 0.2251 &  20 mins   \\ 
    CNN+NN     & 10 &  0.1323 & 0.2985 & 3 hrs    \\ 
    CNN+LSTM & 1    & 0.1001 & 0.4087 &  3 hrs 15 mins   \\ 
    CNN+LSTM     & 10 &  0.1370 & 0.4080 &  7 hrs   \\ 
    Advanced & ALL    & 0.1201 & 0.1605 & 10 hrs 45 mins    \\ 
    \bottomrule
  \end{tabular}
\end{table}

\section{Conclusion and Future Direction}

In this project we used CNN and LSTM to predict acceleration of the waymo's autonomous vehicles. The 11 feature that we extracted from the raw waymo open dataset played an important role in determining the acceleration. Apart from these 11 features, images were also crucial in determining acceleration. Since images carry a lot of data we can't exclude that information. Also, treating this problem as a time series problem helped us to extract information from past frames as well. There were few areas which can be improved upon in future. Firstly, the ground truth acceleration we calculated could be faulty and thus in future we need to ensure that the ground truth acceleration as actually matching the AV's acceleration. Also, fine-tuning the models would be another area that could be worked upon. At last, adding more features to the already calculated feature set could be advantageous in predicting the acceleration.

\subsubsection*{Acknowledgments}

This research work was done under the supervision of Prof. Sharon Di and Rongye Shi.

\section*{References}

\small

[1] Bojarski, M., Del Testa, D., Dworakowski, D., Firner, B., Flepp, B., Goyal, P., Jackel, L., Monfort, M., Muller, U., Zhang, J., Zhang, X., Zhao, J., Zieba, K (2016) End to end learning for self-driving cars, Technical report (2016). \url{http://arxiv.org/abs/1604.07316}

[2] Karpathy, A. and Fei-Fei, L. Deep visual-semantic alignments for generating image descriptions. CoRR, abs/1412.2306, 2014.

\end{document}